\documentclass[sigconf]{acmart}

\usepackage{booktabs} 
\usepackage{amsmath}
\usepackage{wrapfig}
\usepackage{subfigure}
\graphicspath{ {figure/} }
\usepackage{xcolor}
\usepackage{tabularx}

\begin{document}
\title{seq2graph: Discovering Dynamic Dependencies from Multivariate Time Series with Multi-level Attention}

\author{Xuan-Hong Dang, Syed Yousaf Shah and Petros Zerfos}
\affiliation{%
  \institution{IBM Thomas J. Watson Research Center}
  \streetaddress{Yorktown Heights, NY 10598}
}
\email{syshah@us.ibm.com, Xuan-Hong.Dang@ibm.com, pzerfos@us.ibm.com}

%
\renewcommand{\shortauthors}{X.H.Dang et al.}


\begin{abstract}
Discovering temporal lagged and inter-dependencies in multivariate time series data is an important task. However, in many real-world applications, such as commercial cloud management, manufacturing predictive maintenance, and portfolios performance analysis, such dependencies can be non-linear and time-variant, which makes it more challenging to extract such dependencies through traditional methods such as Granger causality or clustering. In this work, we present a novel deep learning model that uses multiple layers of customized gated recurrent units (GRUs) for discovering both time lagged behaviors as well as inter-timeseries dependencies in the form of directed weighted graphs. We introduce a key component of Dual-purpose recurrent neural network that decodes information in the temporal domain to discover lagged dependencies within each time series, and encodes them into a set of vectors which, collected from all component time series, form the informative inputs to discover inter-dependencies. Though the discovery of two types of dependencies are separated at different hierarchical levels, they are tightly connected and jointly trained in an end-to-end manner. With this joint training, learning of one type of dependency immediately impacts the learning of the other one, leading to overall accurate dependencies discovery. We empirically test our model on synthetic time series data in which the exact form of (non-linear) dependencies is known. We also evaluate its performance on two real-world applications, (i) performance monitoring data from a commercial cloud provider, which exhibit highly dynamic, non-linear, and volatile behavior and, (ii) sensor data from a manufacturing plant.  We further show how our approach is able to capture these dependency behaviors via intuitive and interpretable dependency graphs and use them to generate highly accurate forecasts.

\end{abstract}




\maketitle

\section{Introduction}
\label{intro}

\begin{figure}[!h]
 \centering
\includegraphics[width=9cm]{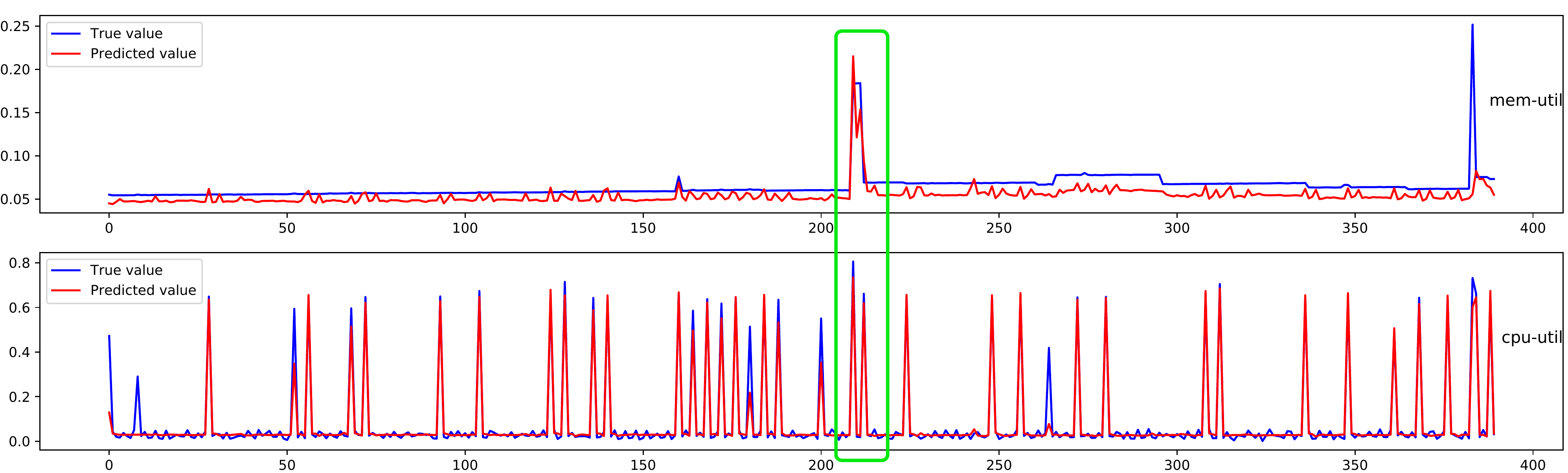}
\caption{cpu- and mem-utility time series from a Cloud Service system. The system behaves dynamically and there is desirable to discover the varying dependencies among component time series along with time, for example, as the ones during the period between 208 and 214 time points  (details are explained in text and further in Experiments section).}
\label{CloudServiceTS}
\end{figure}

Multivariate time series (MTS) modeling and understanding is an important task in machine learning, with numerous applications ranging from science and manufacturing~\cite{laptev2017,shah17}, economics and finance~\cite{Matt17}, to medical diagnosis and monitoring~\cite{gao2017complex}. The main objective of time series modeling is to choose an appropriate model and train it based on the observed data such that it can capture the inherent structure and behavior of the time series. While majority of studies in the literature focus on the task of time series forecasting~\cite{Brockwell16}, in this work we focus on an another equally important and challenging task of discovering dependencies that might exist in the time-variant multivariate times series. This task is important and particularly useful in a variety of applications such as early identification of components that require maintenance servicing in manufacturing facilities~\cite{kdd14predMaintenance}, predicting causalities for resource management in cloud service applications~\cite{shah17}, or discovering leading performance indicators for financial analysis of stocks. In the literature, techniques such as learning coefficients from vector auto regression (VAR) models~\cite{kilian2017}, Granger causality~\cite{eichler2007granger} along with the variants of graphical Granger~\cite{ACML2017}, clustering~\cite{cloudscoutTPDPS17}, or the recently proposed analysis of the weights embedded in a trained recurrent neural network ~\cite{lai2017} can be used to explore inter-dependencies among time series. However, one of the key drawbacks of all these methods is that, once the models have been trained, their learnt coefficients/weights stay \textit{fixed} during the \textit{inference phase}, and hence intrinsically lack the capability of timely discovery of dynamic dependencies among time series, which are particularly important to the aforementioned real-world applications. Figure~\ref{CloudServiceTS} illustrates an example of cloud service with the plot of two component time series, the cpu-utilization and memory-utilization. Between time points of 208th and 214th (in green rectangle), the time series behave dynamically. There is desirable to early predict and understand what cause dynamic change(s) and how have the dependencies among time series varied during this period of time.

To explore time variant dependencies, we introduce a novel deep learning architecture, which is built upon the core of multi-layer customized recurrent neural networks (RNNs). Our model is capable of timely discovering the inter-dependencies and temporal-lagged dependencies that exist in multivariate time series data and represent them as directed weighted graphs. By means of varied inter-dependencies, our model aims at learning and discovering the (unknown) relationships existed among the time series, especially such mutual relationship can vary with time as the generating process of the multivariate series evolves. By varied temporal dependency, our model aims at discovering the time-lagged dependency within each individual time series, to identify those historical data points that highly influence the future values of the multivariate time series. Our proposed model is a multi-layer RNNs network developed based on the success of recently introduced gated recurrent units (GRUs)~\cite{cho2014properties} which excel in capturing long term dependencies in the sequential data, are less susceptible to the present of noise, and readily learn both linear and non-linear relationships embedded in the time series. Our model extracts hidden lagged and inter-dependencies in the MTS and present them in form of dependencies graphs that help to comprehend the time series behaviors along the temporal dimension. In particular, we make the following contributions in the paper: (1) To our best knowledge, our work is the first that addresses the challenging and important problem of discovering dynamic temporal-lagged and inter-dependencies from a system generating multivariate time series during the inference phase in a timely manner with the RNNs deep learning approach. (2) We present a novel neural architecture that is capable of discovering \textit{non-linear} and \textit{non-stationary} dependencies in the multivariate time series, which are not captured by the Granger causality. Of particular importance is its capability of identifying non-stationary, time-varying dependencies during the \textit{inference} phase of the time series, which differentiates our method from state-of-the-art techniques. (3) We evaluate the performance of our proposed approach on synthetic data, for which we have full control and knowledge of the (non-linear) dependencies that exist therein, as well as two real-world datasets, the utilization measures of a commercial cloud service provider, and the sensor data from a semiconductor manufacturing facility, which exhibit highly dynamic and volatile behavior. (4) We compare our model with well-established multivariate time series algorithms, including the widely used statistical VAR model~\cite{kilian2017}, the recently developed RNNs~\cite{salehinejad2017recent} and the hybrid residual based RNNs~\cite{goel2017r2n2}, showing its superior in discovering dependencies
that are also not captured by the Granger causality~\cite{eichler2007granger}.

\vspace*{-0.1cm}
\section{Related work}
\label{related_work}


State-of-the-art studies on time series analysis and modeling can generally be divided into three categories. The first one involves statistical models such as the autoregressive integrated moving average \textit{(ARIMA)}~\cite{Brockwell16}, or the latent space based Kalman Filters~\cite{takeda2016using}, the HoltWinters technique~\cite{tratar2016comparison}, and the vector autoregression (VARs)~\cite{Brockwell16} which are extended to deal with multivariate time series. Although having been the  widely applied tools, the usage of these models often requires the strict assumption regarding the linear dependencies of the time series. Their approximation to complex real-world problems does not always lead to satisfactory results~\cite{hsu2017time}. The second category consists of techniques based on artificial neural networks (ANN)~\cite{chen2015comparative}. One of the earliest attempts~\cite{chakraborty1992forecasting} employs the 1-hidden feedforward neural network with markedly improved performance as compared to the ARIMIA. 
More recent approaches include the usage of multilayered perceptron network~\cite{choubin2016multiple}, Elman recurrent neural network~\cite{ardalani2010chaotic}, and the RNNs dealing classification task over short sequences~\cite{raj18,choi2016retain}.
Compared to statistical models, ANN-based models have a clear advantage of well approximating non-linear complex functions. However, they are also criticized for their black box nature and limited explanatory information. 
The third category is a class of hybrid methods that combine several models into a single one. \cite{zhang03} is among the premier studies that first applies the ARIMA to model the linear part of time series, and then adopts a feedforward neural network to model the data residuals returned by the ARIMA. Subsequent studies are often the variants of this approach, by replacing the feedforward network with other learning models~\cite{khashei2011novel}, such as the long-short term memory (LSTM)~\cite{goel2017r2n2}, or non-linear learning techniques~\cite{hu2013hybrid}. These existing techniques including Granger causality~\cite{eichler2007granger,ACML2017} either lack the ability to discover non-linear dependencies or are unable to capture the instantaneous dynamic dependencies in the data, as contrasted to our studies that learn and discover these time-variant and complex dependencies explicitly through the deep recurrent neural network model.

\vspace*{-0.2cm}
\section{Methodology}
\label{Methodology}

This paper focuses on a multivariate time series (MTS) system with $D$ time series (variables) of length $T$, denoted by $X = \{\mathbf{x}_1, \mathbf{x}_2,\ldots,\mathbf{x}_T\}$, where each $\mathbf{x}_t \in \mathcal{R}^D$ is the measurements of MTS at timestamp $t$. The $d$-th time series is denoted by $X^d = \{x_1^d,x_2^d,\ldots,x_T^d\}$, in which $x_t^d\in \mathcal{R}$ represents its measurement at time $t$.
Given such an MTS system, we aim to analyze it from two aspects: (i) how the future measurements are dependent on the past values of each individual time series, we call them temporal lagged dependencies; (ii) how the measurements of different time series are dependent on each other, we call them inter-timeseries dependencies. The inter-dependencies discovered at a specific timestamp can be intuitively represented as a directed graph whose nodes correspond to individual time series, and an edge reveals which time series is dependent on which time series. It is important to note that our focus is to capture the \textit{time-variant} dependencies not limited to the training phase but also during the inference phase, thus timely discovering the dynamically changing dependencies at any timestamp $t$, and utilize these unearthed dependencies to accurately forecast future values of the MTS.

\vspace*{-0.2cm}
\subsection{Overall model}

 \begin{figure}[t]
 \centering
 \includegraphics[width=1.1\linewidth]{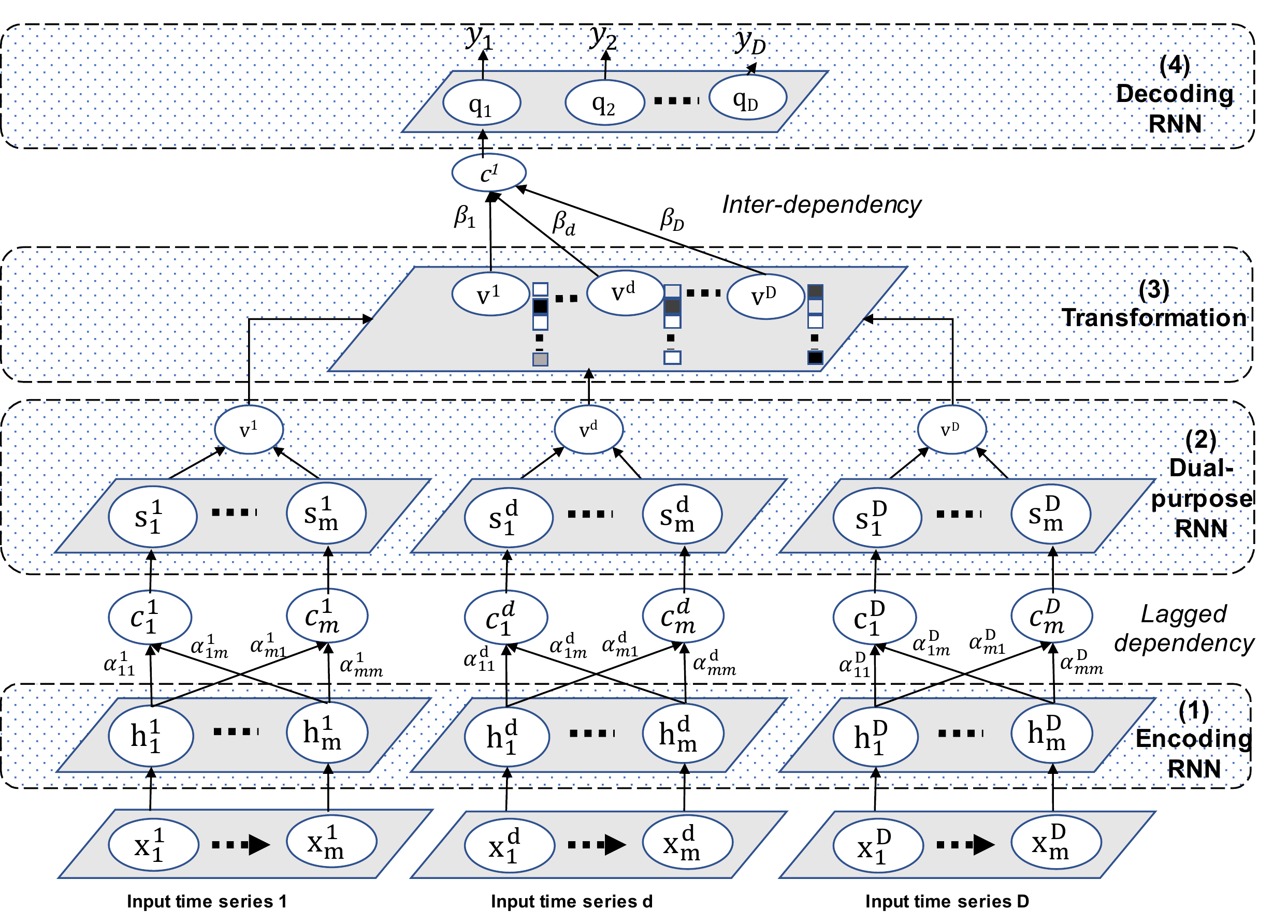}
 \caption{Overview of our deep learning model. Given an MTS at input, our model learns to discover temporal lagged and inter-dependencies, which are outputted as dependency graphs while making prediction over the future time point $\mathbf{y} $.}
 \vspace*{-0.5cm}
 \label{architecture}
 \end{figure}




We outline our network model in Figure \ref{architecture} which consists of 4 main components/layers: (1) Given a multivariate time series at input layer, the  \verb"Encoding RNN" component comprises a set of RNNs, each processes an individual time series by encoding them into a sequence of encoding vectors. (2) The next  \verb"Dual-purpose RNN" component also consists of a set of customized RNNs, each discovers the temporal lagged dependencies from one constituted time series and subsequently encodes them to a sequence of output states. (3) Sequences of output states from all component Dual-purpose RNNs in the previous layer are gathered together and each is transformed into a high dimensional vector by the  \verb"Transformation layer". These feature vectors act as the encoding representations of the constituted time series prior to the next level of identifying inter-dependencies among series. (4) The final \verb"Decoding RNN" discovers the inter-dependencies among all component time series through identifying the most informative input high dimensional vectors toward forecasting the next value of each time series at the final output of the model. This final component hence outputs both the dependency graphs and the predictive next value for all component time series. 
Before presenting each component/layer of the model in details, we provide a brief background on the gated recurrent units~\cite{cho2014learning} to which our components are extension variants intrinsically designed for the multivariate time series setting.

\noindent \textit{Gated recurrent units (GRUs)}: have been among the most successful neural networks in learning from sequential data~\cite{cho2014learning,salehinejad2017recent}. Similar to the long-short term memory units (LSTMs)~\cite{hochreiter1997long}, GRUs are cable of capturing the long term dependencies in sequences through the memory cell mechanism. Specifically, there are two gates, reset $\mathbf{r}_t$ and update $\mathbf{z}_t$,  defined in GRUs which are computed with the following equations.

\vspace*{-0.6cm}
\begin{align}
\label{eq:rt} 
\hspace*{-0.5cm}
\mathbf{r}_t = \sigma(\mathbf{W}_r\mathbf{x}_t + \mathbf{U}_r \mathbf{h}_{t-1} + \mathbf{b}_r)
\end{align}

\vspace*{-0.6cm}

\begin{align} 
\label{eq:zt} 
\hspace*{-0.5cm}
\mathbf{z}_t = \sigma(\mathbf{W}_z\mathbf{x}_t + \mathbf{U}_z \mathbf{h}_{t-1} + \mathbf{b}_z) 
\end{align}

%

%

\noindent with $\sigma(.)$ as the non-linear sigmoid function, $\mathbf{x}_t$ is the data point input at time $t$, $\mathbf{h}_{t-1}$ is the previous hidden state of the GRUs. $\mathbf{W}$'s,$\mathbf{U}$'s, and $\mathbf{b}$'s (subscripts are omitted) are respectively the input weight matrices, recurrent weight matrices, and the bias vectors, which are the GRUs' parameters to be learnt. Based upon these two gates, GRUs compute the current memory content $\tilde{\mathbf{h}}_t $ and then update their hidden state (memory) to the new one $\mathbf{h}_t $ as follows:

\vspace*{-0.4cm}
\begin{align} 
\label{eq:htbar} 
\tilde{\mathbf{h}}_t = \tanh(\mathbf{W}_hx_t + \mathbf{U}_h (\mathbf{r}_t\odot \mathbf{h}_{t-1}) + \mathbf{b}_h)
\end{align}
\vspace*{-0.65cm}

\vspace*{-0.4cm}
\begin{align} 
\label{eq:ht} 
\mathbf{h}_t = (1-\mathbf{z}_t)\odot \mathbf{h}_{t-1} + \mathbf{z}_t\odot \tilde{\mathbf{h}}_t\end{align}

\noindent with $\odot$ as the element-wise product, and  $\mathbf{W}_h, \mathbf{U}_h, \mathbf{b}_h$ are again the GRU's parameters. As seen, $\tilde{\mathbf{h}}_t $ is determined by the new data input $x_t$ and the past memory $\mathbf{h}_{t-1}$, and its content directly impacts $\mathbf{h}_{t}$ through Eq.\eqref{eq:ht}. It is due to the usage of $\mathbf{r}_t$ and $\mathbf{z}_t$ in these equations that defines their specific functions (despite their similar formulae in Eqs.\eqref{eq:rt},\eqref{eq:zt}). $\mathbf{r}_t$ decides how much information in the past $\mathbf{h}_{t-1}$ should be forgotten (Eq.\eqref{eq:htbar}, hence named reset gate), while $\mathbf{z}_t$ determines how much of the past information to pass along to the current state (Eq.\eqref{eq:ht}, hence named update gate). Due to this gating mechanism, GRUs can effectively keep most relevant information at every single step. In this paper, we denote all above computational steps briefly by $\mathbf{h}_t = GRU(\mathbf{x}_t,\mathbf{h}_{t-1})$ implying it as a function with inputs and outputs (skipping the internal gating computations).  

\vspace*{-0.2cm}
\subsection{Discovering temporal lagged dependencies}

At a single time point, our model receives $D$ sequences as inputs, each consisting of $m$ historical values from each component time series, and produces at output a graph of inter-dependencies among all time series along with a sequence $\mathbf{y} = \{y_1,\ldots,y_D\}$ as the forecasted values of the next timestamp in the MTS (Fig.\ref{architecture}). While being trained to map the set of input sequences to the output sequence $\mathbf{y}$, our model learns to discover temporal lagged dependencies within each time series at the low layers, and the inter-dependencies among all time series at the higher layer of the model. It is important to note that these dependencies can be complex, non-linear, and especially time-variant. For instance, future performance of CPUs and memory usage from a cloud management service can be anticipated based on their own historical values in most normal situations. Nonetheless, when the requesting data from outside world are abnormally enormous (e.g., at peak hours, or under attacks), overwhelming the CPUs' capacity, poor performance on CPUs can be experienced and consequently impact future performance of memory due to caching data issues. Early detection and accurately discovery these fast changing dependencies in an online fashion is particularly important since it can enable an automate system, or at least support the cloud administrator to act timely and appropriately, ensuring the sustainability of the cloud service. Our deep learning model is intrinsically designed for modeling and discovering these complex time-varying dependencies among time series. We next present the network components designed for discovering the time lagged dependencies which tells us what timepoints in the past of a specific $d$-th time series that crucially impact the future performance of MTS. 

\noindent \textit{(1) Encoding RNN:}  At the $d$-th time series, the Encoding RNN component receives a sequence of the last $m$ historical values $\{x^d_1,x^d_2,\ldots,x^d_m\}$, and it encodes them into a sequence of hidden states $\{\mathbf{h}^d_1,\mathbf{h}^d_2,\ldots,\mathbf{h}^d_m\}$. Its main objective is to learn and encode the long-term dependencies among historical timepoints of the given time series into a set of the RNNs' hidden states. As our model read in an entire sequence of length $m$ timepoints from the $d$-th time series, we utilize the bidirectional $GRU_E$ ($E$ for encoding) in order to better exploit information from both directions of the sequence, as shown by the following equation, for each $t=1\ldots m$:

\vspace*{-0.4cm}
\begin{align*}
\mathbf{h}^d_t = [\overrightarrow{\mathbf{h}^d_t},\overleftarrow{\mathbf{h}^d_t}]=[\overrightarrow{GRU_E}(x^d_t,\overrightarrow{\mathbf{h}^d_{t-1}}),\overleftarrow{GRU_E}(x^d_t,\overleftarrow{\mathbf{h}^d_{t+1}})] 
\end{align*}
\vspace*{-0.3cm}

\noindent \textit{(2) Dual-purpose RNN:}  At this component, there is a corresponding variant $GRU_{DP}$ network for each $GRU_E$ in the previous Encoding RNN layer. Unlike a $GRU_E$ which outputs its hidden states, $GRU_{DP}$ explicitly returns $m$ vectors $v^d_{t}$'s based on $\mathbf{h}^d_t$, its own hidden states $\mathbf{s}^d_t$, and the novelly introduced temporal context vectors $\mathbf{c}^d_t$'s. In calculating each $\mathbf{c}^d_t$, $GRU_{DP}$ generates non-negative, normalized coefficients $\alpha^d_{tj}$ w.r.t. each input vector $\mathbf{h}^d_j$ from the $GRU_E$. While still retaining the time order among the output values, this mechanism enables our network component to focus on specific timestamps of the $d$-th time series at which the most relevant information is located. Mathematically, our $GRU_{DP}$ network computes the following Eqs.\eqref{eq:at}-\eqref{eq:vd}, for each $t = 1,\ldots, m$:

%
%

\vspace*{-0.2cm}
\begin{align} 
\label{eq:at}
 \alpha^d_{tj} = \frac{\exp(tanh(\mathbf{W}_\alpha[\mathbf{s}^d_{t-1};\mathbf{h}^d_j])^\top \mathbf{u}_\alpha)}{\sum_{k=1}^m \exp(tanh(\mathbf{W}_\alpha[\mathbf{s}^d_{t-1};\mathbf{h}^d_k])^\top \mathbf{u}_\alpha)} 
\end{align}

\vspace*{-0.2cm}
\begin{align}
\label{eq:vt} 
\mathbf{c}^d_t = \sum_{j=1}^m \alpha^d_{tj}\mathbf{h}^d_j \qquad \text{} 
\end{align}

\vspace*{-0.2cm}
\begin{align}
\label{eq:st}
\mathbf{s}^d_t = GRU_{DP}(v^d_{t-1},\mathbf{s}^d_{t-1}, \mathbf{c}^d_t)
\end{align}

\vspace*{-0.3cm}
\begin{align} 
\label{eq:vd} 
v^d_{t}  = tanh(\mathbf{W}^d_ov^d_{t-1} + \mathbf{U}^d_o\mathbf{s}^d_t  + \mathbf{C}^d_o  \mathbf{c}^d_t + \mathbf{b}^d_o) 
\end{align}
\vspace*{-0.3cm}

\noindent where $\mathbf{W}_\alpha$ and $\mathbf{u}_\alpha$ are parameters to learn $\alpha^d_{tj}$'s coefficients, and they are jointly trained with the entire model. $\mathbf{W}^d_o,  \mathbf{U}^d_o, \mathbf{C}^d_o, \mathbf{b}^d_o$ are the network's parameters in learning the outputs $v^d_{t}$'s. As seen, $\alpha^d_{tj}$'s are utilized to weight the importance of each encoding vector $\mathbf{h}^d_j$ toward learning the temporal context vector $\mathbf{c}^d_t$, which directly makes impact on updating the current hidden state $\mathbf{s}^d_t$ and the output  $v^d_{t}$. It is worth mentioning that our weight mechanism through $\alpha^d_{tj}$'s at this temporal domain follows the general attention mechanism adopted in neural machine translation (NMT)~\cite{bahdanau2014neural,luong2015effective}, yet it is fundamentally different in two aspects. First,  we do not have ground-truth (e.g., target sentences in NMT) for the outputs $v^d_{t}$'s but let $GRU_{DP}$ learn them automatically. Their embedding information directly governs the process of learning inter-timeseries dependencies in the upper layers in Fig.\ref{architecture}. Indeed, $v^d_{t}$'s act as the bridging information between two levels of discovering the time lagged and the inter-dependencies. Second, our model explores the $tanh$ function rather than the $softmax$ one, which gives it more flexility to work on the \textit{continuous} domain of time series, making real values embedded in $v^d_{t}$'s, similar to the input time series. Hence, this extension of GRU variant essentially performs two tasks: (i) decoding information in the temporal domain to discover time lagged dependencies within each time series, (ii) encoding this temporal information into a set of outputs $v^d_{t}$'s which, collected from all time series, form the inputs for the next layer of discovering inter-dependencies among all time series. For this reason, we name this layer the Dual-purpose RNN.

\subsection{Discovering inter-dependencies}

\noindent \textit{(3) Transformation layer:}  Following the Dual-purpose RNN layer, which generates a  sequence $\mathbf{v}^d =\{v^d_1,v^d_2,\ldots,v^d_m\}$ for $d$-th time series, this Transformation layer gathers these sequences from all $D$ time series and subsequently transforms each into a feature vector. These  high dimensional vectors are stacked to a sequence, denoted by $\{\mathbf{v}^1, \mathbf{v}^2, \ldots, \mathbf{v}^D\}$, and serves as the input for the Decoding RNN component. As noted, though there is no specific temporal dimension among these vectors, their order in the stacked sequence must be specified prior to the model training to ensure the right interpretation when our model explores the inter-timeseries dependencies.

\noindent \textit{(4) Decoding RNN:}  This component consists of a single variant $GRU_D$ ($D$ for decoding) network that performs the inter-dependencies discovery while also making prediction over $\mathbf{y}$ at the model's output. During the training phase, an $y_i\in\mathbf{y}$ is the next value of the $i$-th time series, and its computation is relied on all vectors $\{\mathbf{v}^1, \mathbf{v}^2, \ldots, \mathbf{v}^D\}$ released by the previous Dual-purpose RNN layer. Its calculation steps are as follows:

%
%

\vspace*{-0.3cm}
\begin{align} 
\label{eq:bd}
\beta^d_{i} = \frac{\exp(tanh(\mathbf{W}_\beta[\mathbf{q}_{i-1};\mathbf{v}^d])^\top \mathbf{u}_\beta))}{\sum_{k=1}^D \exp(tanh(\mathbf{W}_\beta[\mathbf{q}_{i-1};\mathbf{v}^k])^\top \mathbf{u}_\beta))} 
\end{align}

\vspace*{-0.4cm}
\begin{align} 
\label{eq:ci}
\mathbf{c}_i = \sum_{d=1}^D \beta^d_{i} \mathbf{v}^d
\end{align}

\vspace*{-0.4cm}
\begin{align}
\label{eq:qi}
\mathbf{q}_i = GRU_D(\mathbf{q}_{i-1}, \mathbf{c}_i)
\end{align}

\vspace*{-0.4cm}
\begin{align} 
\label{eq:yi}
y_i  = tanh(\mathbf{C}_o  \mathbf{c}_i + \mathbf{U}_o\mathbf{q}_i + \mathbf{b}_o) 
\end{align}
 \vspace*{-0.3cm}

\noindent where $\mathbf{W}_\beta$ and $\mathbf{u}_\beta$ are parameters to learn $\beta^d_{i}$'s weights, and $\mathbf{C}_o, \mathbf{U}_o$ and $\mathbf{b}_o$ are the parameters w.r.t. the output $y_i$'s. As seen, this network component first computes the alignment of each of vectors $\mathbf{v}^d$ (featured for each input time series at this stage) with the hidden state $\mathbf{q}_{i-1}$ of the $GRU_D$ through the coefficient $\beta^d_{i}$. Using these generated normalized coefficients, it obtains the context vector $\mathbf{c}_i$ that is used to update the current $GRU_D$'s hidden state $\mathbf{q}_{i}$ and making prediction of output $y_i$. Let $y_i$ be the next value of the $i$-th time series, $\beta^d_i $ will tell us how important the corresponding $d$-th time series (represented by $\mathbf{v}^d$) in predicting $y_i$. In other words, it captures the dependency of  $i$-th series on the $d$-th series. The stronger this dependency is, the closer to 1 the $\beta^d_i$. The whole vector $\mathbf{\beta}_i$ therefore tells us the dependencies of $i$-th time series on all $D$ time series in the system. These coefficients along with the predictive values are the outputs of our entire model, and based on $\beta^d_i$'s, it is straightforward to construct the graph of inter-dependencies among all time series. 

\vspace*{0.1cm}
\textit{Optimization:} We train our entire model end-to-end via the adaptive moment estimation algorithm~\cite{adam} using mean squared errors as the train loss function. Let us denote the entire model as a function $F$ with $\theta$ as its all parameters, the train loss function is:

\begin{align} 
\label{eq:loss}
L(X;\theta)  = \sum_{t=m}^{T-1}\sum_{d=1}^D(y_d - F_d(X_t;\theta))^2
\end{align}

\noindent with $X$ as the entire training MTS data, $X_t$ as a sequence of the last $m$ timepoints in MTS at timestamp $t$, $y_d$ as the value of $d$-th time series at the next timestamp $t+1$ and $F_d(X_t;\theta)$ as the predictive value of the model toward $y_d$.

\begin{figure*}[!tb]
\includegraphics[width=18cm]{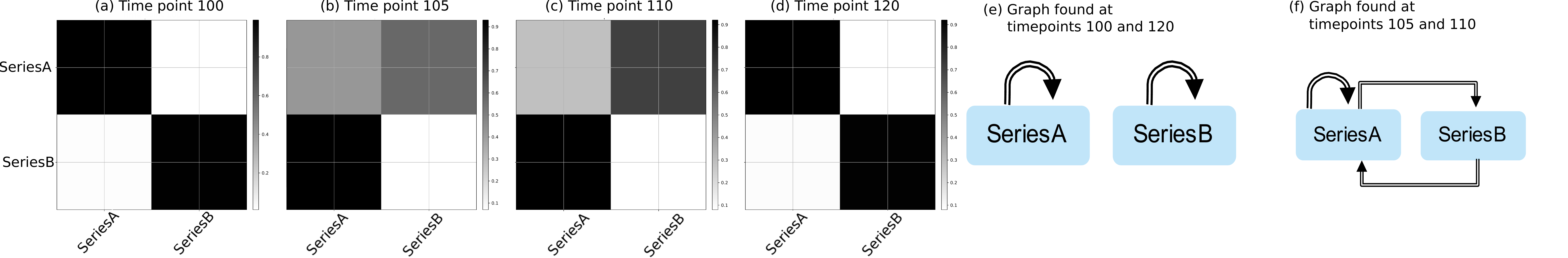}
\vspace*{-0.8cm}
\caption{Inter-dependencies discovered by seq2graph in the bivariate time series. In (a)-(d), the darker shows the more relevant of series shown in column in predicting series shown in row. (e)-(f) show the corresponding graphs derived from these coefficients.}
\vspace*{-0.5cm}
\label{Syn-SA}
\end{figure*}

%
%
\paragraph{Discussion:} Our model though explores the temporal lagged and inter-dependencies among time series, it can also be more generally seen as performing the task of transforming multiple input sequences into one output sequence. As presented above, the output sequence is the values of the next timestamp of all time series constituted in MTS, but one can easily replace this sequence with the next $n$ values of \textit{one} time series of interest. That means, we want to discover dependencies of this particular series on the other ones while forecasting its multiple time points ahead. With this case, Eq.\eqref{eq:qi} and \eqref{eq:yi} can be replaced by adding a term accounting for the previous predictive value, i.e., $\mathbf{q}_i = GRU_D(y_{i-1},\mathbf{q}_{i-1}, \mathbf{c}_i)$ and $y_i  = tanh(y_{i-1},\mathbf{C}_o  \mathbf{c}_i + \mathbf{U}_o\mathbf{q}_i + \mathbf{b}_o)$ in order to further exploit the temporal order in the output sequence. Interpretation over the inter-dependencies based on $\mathbf{\beta}_i$'s remains unchanged yet it is for this solely interested time series and over a window of its $n$ future time points.

\section{Experiments}
\label{experiment}

We name our model \verb"seq2graph" and its training were performed on a machine with two Tesla K80 GPUs, running Tensorflow 1.3.0 as backend. We compare performance of \verb"seq2graph" against: (i) \verb"VAR" model~\cite{kilian2017} which is one of the most widely used statistical techniques for analyzing MTS; (ii) \verb"RNN-vanilla" which receives all time series at inputs and predicts $\mathbf{y}$ at output (this model can be considered as an ablation study of our network by collapsing all layers into a single one). (iii) \verb"RNN-residual"~\cite{goel2017r2n2}, a hybrid model that trains an RNN on the residuals returned by the statistical ARIMA. To match our model complexity, we use GRUs for both \verb"RNN-vanilla" and \verb"RNN-residual" with similar number of hidden neurons used in our model, and their number of stacked layers are tuned from one to three layers. For \verb"seq2graph", its optimal number of hidden neurons for each network component is tuned from the pool of $\{16,32,48,64\}$. For each examined dataset presented below, we split it into training, development, and test sets with the ratio of 70:15:15. The development set is used to tune models' hyper-parameters while the test set is used to report their performance.


\vspace*{-0.2cm}
\subsection{Empirical analysis on synthetic data}
Discovering time lagged and inter-dependencies is challenging and there is no  available dataset with known ground truths. Hence, we make use of synthetic data and attempt to answer the following key questions: (i) Can \verb"seq2graph" discover nonlinear time-varying dependencies (e.g., introduced in data via \textit{if-then} rules) among time series during inference phase? (ii) Once dependencies of one time series on the others are known, can \verb"seq2graph" accurately discover the lagged dependencies? (iii) How accurate is our model in predicting future values of the multivariate time series?

\textit{Data description:} For illustration purposes, we use the bivariate time series (MTS with more variables are evaluated on the real datasets) that simulates the \textit{time-varying} interaction between two time series namely SeriesA and SeriesB. We generate 10,000 data points with real-values between 0 and 1 using following two rules:


\begin{align*} 
\small
\hspace*{-.40cm}
 \{A[t\!\!+\!\!1],B[t\!\!+\!\!1]\}\!\! = \!\!
\begin{cases}
 \{A[t\!\!-\!\!4], f(B[t\!\!-\!\!3],B[t\!\!-\!\!6])\}
   \text{if~~} A[t]\!\!<\!\!0.5  \\
 \{f(A[t\!\!-\!\!6],B[t\!\!-\!\!3]), A[t\!\!-\!\!3]\}  
   \text{if~~} A[t]\!\! \geq\!\! 0.5
  \end{cases}
\end{align*}
\normalsize

\noindent Rule 1 (1st row) specifies that, if the current value of SeriesA is smaller than 0.5, values in the next future timestamp $t+1$ of both series are autoregressive from themselves.  Rule 2 (2nd row) specifies that if this value is $\geq 0.5$, there are mutual impacts between two series. Specifically, the next value of SeriesA is decided by its value at the 6th lagged timestamp and that of the 3rd lagged in SeriesB, while the next value of SeriesB is determined by the 3rd lagged time point in SeriesA. $f$ is the average function over these lagged values. Also, in order to ensure the randomness and volatility in the input sequence, we frequently re-generated values for the lagged time points. 


\paragraph{Discovery of dependencies:} We trained all models with window size $m=8$, just above the furthest lagged points defined in our data generating rules. We evaluate the capability of \verb"seq2graph" in discovering dependencies through investigating the time-varying vectors $\mathbf{\alpha}$'s and  $\mathbf{\beta}$'s during the inference phase (i.e. on test data). Recall that, unlike model's parameters which were fixed after the training phase, these coefficient vectors keep changing according to the input sequences. We show in Fig.\ref{Syn-SA}(a)-(d) the coefficients of $\mathbf{\beta}$'s (Eq.\eqref{eq:bd}) which reflects the inter-dependencies between two time series at 4 probed timestamps on the test data. Within each figure, a darker color shows a stronger dependency of time series named in row on those named in column (i.e., each row in the figure encodes a vector $\mathbf{\beta}_i$ whose entries are computed by Eq.\eqref{eq:bd}). 

Investigating the input sequences, we knew that data points at timestamps 100 and 120 were generated by rule 1, while those at 105 and 110 were generated by rule 2. It is clearly seen that \verb"seq2graph" has accurately discovered the ground truth inter-dependencies between SeriesA and SeriesB. Particularly, at timestamps 100 and 120 at which the two time series were generated by rule 1, our model found that each time series auto-regresses from itself, as evidenced by the dark color at the diagonal entries in Figs.\ref{Syn-SA}(a) and (d). However, at timestamps 105 and 110, \verb"seq2graph" found that SeriesA's next value is dependent on the historical values of both SeriesA and B, while SeriesB's next value is determined by the historical values of SeriesA (detailed by the lagged dependencies shortly shown below), as observed from  Figs.\ref{Syn-SA}(b) and (c). These inter-dependencies between the two time series can be intuitively visualized via a directed weighed graph generated at each time point. In Fig.\ref{Syn-SA}(e), we show the typical graph found at timestamps 100 and 120, while in Fig.\ref{Syn-SA}(f). we plot the one found at timepoints 105 and 110. As seen, these directed graphs clearly match our ground truth rules that have generated the bivariate time series.

\begin{figure}[!t]
\hspace*{-0.5cm}
\includegraphics[width=10cm]{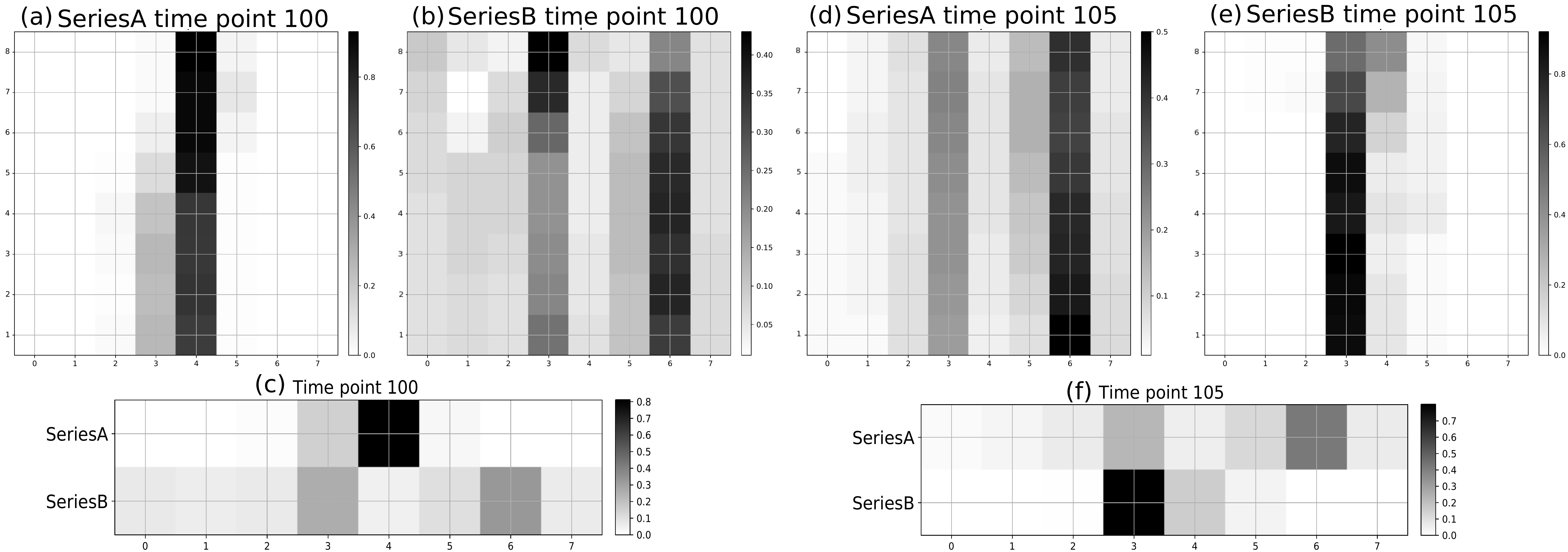}
\vspace*{-0.5cm}
\caption{Temporal lagged dependencies discovered by our model at 2 time points 100 ((a)-(b)) and 105 ((d)-(e)). Plots (c) and (f) are respectively the row-aggregate versions of Plots (a)-(b) and (d)-(e) (detailed interpretation is given in text).}
\label{Syn-TA}
\end{figure}

In evaluating the correctness of temporal lagged dependencies discovered by \verb"seq2graph", we select 2 time points 100 and 105 (each represented for one generated rule), and further inspect the vectors $\mathbf{\alpha}$'s (Eq.\eqref{eq:at}) calculated at the Dual-purpose RNN layer in our deep neural model. Figs.\ref{Syn-TA}(a)-(b) show coefficients of these vectors w.r.t. to input SeriesA and B respectively at time points 100, while Figs.\ref{Syn-TA}(a)-(b) show those at time point 105. Within each figure, the x-axis shows the lagged timestamps with the latest one indexed by 0, while the y-axis shows the index of $t$ computed in the Dual-purpose RNN. Again, dark colors encode dependencies on the last $m$ lagged timepoints in a time series. One can observe that \verb"seq2graph" properly puts more weights (larger coefficients $ \alpha^d_{tj}$'s) to the correct lagged time points in both cases. Particularly, it strongly focuses on 4th lagged point in SeriesA, 3rd and 6th lagged points in SeriesB (Figs.\ref{Syn-TA}(a)-(b)) at timestamp 100, and gives much attention to 3rd and 6th lagged points in SeriesA, and 3rd lagged point in SeriesB at time 105 (Fig.\ref{Syn-TA}(d)-(e)). Note that, when the number of time series in a system is large (as we shortly present with the real-world time series systems), one can aggregate $\mathbf{\alpha}$'s w.r.t each component time series into a single one for better visualization. For example, 
Figs.\ref{Syn-TA}(a) and (b) are respectively row-aggregated into the first and second rows of Fig.\ref{Syn-TA}(c). Likewise, Fig.\ref{Syn-TA}(d) and (e) are aggregated into the two rows of Fig.\ref{Syn-TA}(f). As seen, both figures explain well the lagged dependencies within each component time series (named in the y-axis) and they match our ground truth rules. 

In comparison, we found that these \textit{dynamically} changed dependencies were not accurately captured by \verb"VAR"'s and \verb"RNN-residual"'s coefficients. They both tend to give weights on more historical time points than needed. For instance, \verb"VAR" places high (absolute) weights on timepoints $\{1,3,6,7,8\}$ on SeriesA, and $\{3,5,6\}$ on SeriesB when it was trained to forecast the future value of SeriesA; on timepoints $\{1,3,5,7\}$ on SeriesA and $\{3,4,5,7\}$ on SeriesB when it was trained to forecasts the next value of SeriesB. None of algorithms are able to timely discover \textit{time-varying} dependencies in the MTS.

\paragraph{Prediction Accuracy:}We briefly provide prediction accuracy of \verb"seq2graph" and other techniques on this dataset in the first two columns of Table~\ref{tbl-prediction-accuracy}. The root mean squared error (RMSE), and mean absolute error (MAE) have been used for the accuracy measurements. As observed, \verb"VAR" underperforms compared to both \verb"RNN-vanilla" and \verb"RNN-residual", which is due to its linear nature. The prediction accuracy of \verb"seq2graph" is better than \verb"RNN-vanilla" and \verb"RNN-residual" while further offering insights into the dependencies within and between time series. Its better performance could be justified by the accurate capturing of time-varying time series dependencies. 


\vspace*{-0.3cm}

\subsection{Empirical analysis over real-world data sets }

\begin{figure*}[!tb]
\hspace*{-1cm}
\includegraphics[width=20cm]{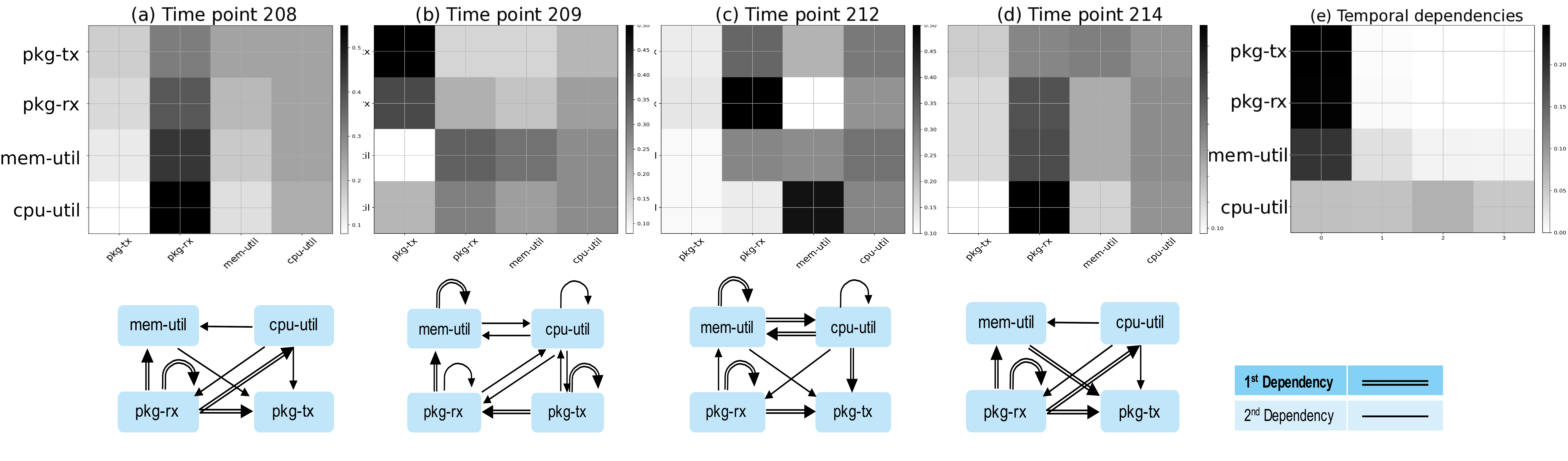}
\vspace*{-0.7cm}
\caption{Cloud Service Data: common inter-dependencies are shown at timepoints 208, 214. They are dynamically changed between timepoints 209 and 212. Their corresponding directed graphs are shown in the second row. Thickness of edges is proportional to strength of dependencies. Common time lagged dependencies are shown in the 5th plot (x-axis shows lagged timestamp with the latest one indexed by 0).}
\label{IBM-cloud-SA}
\end{figure*}

 \begin{table*}[tbp]
  \centering
   \caption{Prediction accuracy of all methods. A better performance means close to zero of RMSE and MAE.}
    \resizebox{\textwidth}{!}{%
    \begin{tabular}{llrrrrrrrrrrr}
    \toprule
          &       & \multicolumn{2}{c}{Synthetic Data} & \multicolumn{4}{c}{Cloud Service Data} & \multicolumn{5}{c}{Manufacturing Data} \\
\cmidrule{3-13}          &       & \multicolumn{1}{l}{SeriesA} & \multicolumn{1}{l}{SeriesB} & \multicolumn{1}{l}{pkg-tx} & \multicolumn{1}{l}{pkg-rx} & \multicolumn{1}{l}{mem-util} & \multicolumn{1}{l}{cpu-util} & \multicolumn{1}{l}{CurrentHTI} & \multicolumn{1}{l}{PowerHTI} & \multicolumn{1}{l}{PowerSPHTI} & \multicolumn{1}{l}{VoltHTI} & \multicolumn{1}{l}{TempHTI} \\
    \midrule
    RMSE  & VAR   & 0.22  & 0.215 & 0.04  & 0.024 & 0.021 & 0.053 & 0.022  & 0.028 & 0.029 & 0.019 & 0.012 \\
          & RNN-vanilla & 0.017 & 0.014 & 0.039 & 0.023 & 0.018 & 0.049 & 0.021 & 0.029 & 0.027 & 0.021 & 0.011 \\
          & RNN-residual & 0.026 & 0.025 & 0.038 & 0.021 & 0.015 & 0.06  & 0.02  & 0.024  & 0.027  & 0.018  & 0.012 \\
          & seq2graph & 0.012 & 0.013 & 0.037 & 0.022 & 0.015 & 0.028 & 0.018 & 0.022 & 0.022 & 0.016 & 0.009 \\
    \midrule
    MAE   & VAR   & 0.184 & 0.178 & 0.024 & 0.009 & 0.007 & 0.044 & 0.006 & 0.007 & 0.006 & 0.009 & 0.005 \\
          & RNN-vanilla & 0.09  & 0.08  & 0.021 & 0.009 & 0.006 & 0.042 & 0.006 & 0.005 & 0.007 & 0.008 & 0.005 \\
          & RNN-residual & 0.11  & 0.12  & 0.021 & 0.006 & 0.007 & 0.044 & 0.005 & 0.006 & 0.005 & 0.008 & 0.005 \\
          & seq2graph & 0.08  & 0.07  & 0.023 & 0.007 & 0.005 & 0.015 & 0.004 & 0.005 & 0.004 & 0.006 & 0.004 \\
    \bottomrule
    \end{tabular}%
    }%
  \label{tbl-prediction-accuracy}
\end{table*}%

\textit{Data description:} We evaluate our model on two real-world systems that generate multivariate time series: (i) Cloud Service Data consisting of 4 metrics CPU utilization (cpu-util), memory utilization (mem-util), network receive (pkg-rx), and network-transmit (pkg-tx), collected every 30 seconds from a public cloud service provider over one day. (ii) Manufacturing Data consisting of 5 sensors measured over CurrentHTI, PowerHTI, VoltHTI, TempHTI, and PowerSPHTI (power set point) at a manufacturing plant with 500,000 timepoints. 
\paragraph{Discovery of dependencies:} With regard to the Cloud Service data, we show in the first four plots of Fig.\ref{IBM-cloud-SA} the time-varying dependencies of one typical short period during the inference phase, between timepoints 208 and 214 as highlighted in the green rectangle in Fig.\ref{CloudServiceTS}\footnote{Recall that the time series shown in Fig.\ref{CloudServiceTS} are during the inference time of the Cloud Service data. We plot the forecasted values by our model in red while the ground truth one in blue.}, at which the system is likely suffered from a large amount of requesting data, impacting the performance on both the cpu- and mem-utilization. As observed, there are clear patterns that can be visibly interpreted. Prior and after this period of time (timepoints 208 and 214 respectively), performance of both cpu-util and mem-util are highly dependent on the pkg-rx (Figs.\ref{IBM-cloud-SA}(a,d)). Nonetheless, when the utilization in the memory time series is notably high (between timepoints 209 and 212), such dependencies significantly change. For instance, performance of cpu-util is determined by almost all series at the early stage of the event (timepoint 209) and lately strongly determined by only the mem-util and itself (timepoint 212). They tend to reflect the situation at which the Cloud system suffers from a large amount of data, being buffered in the memory, and in turns requests intensive performance of the cpu-util\footnote{After this period, the system restores to the common state in which the performance of all time series generally depends on the pkg-rx and cpu-util as shown in Figs.\ref{IBM-cloud-SA}(d).}. The detailed inter-dependencies among all time series are illustrated in the weighted directed graph plotted below each time point in Fig.\ref{IBM-cloud-SA}. Thickness of edges in general is proportional to the strength of dependencies. In the plots, we roughly classify them into 1st and 2nd dependencies with the former one being more significant. The common time lagged dependencies are shown in the 5th plot of Fig.\ref{IBM-cloud-SA}. As mentioned, the x-axis shows the lagged timepoints with the latest one indexed by 0. As seen, except the cpu-util whose historical lagged values often influence the future performance of all time series, the latest values of the other time series tend to impact the system's future performance.

\begin{figure*}[!tb]
\hspace*{-1cm}
\includegraphics[width=20cm]{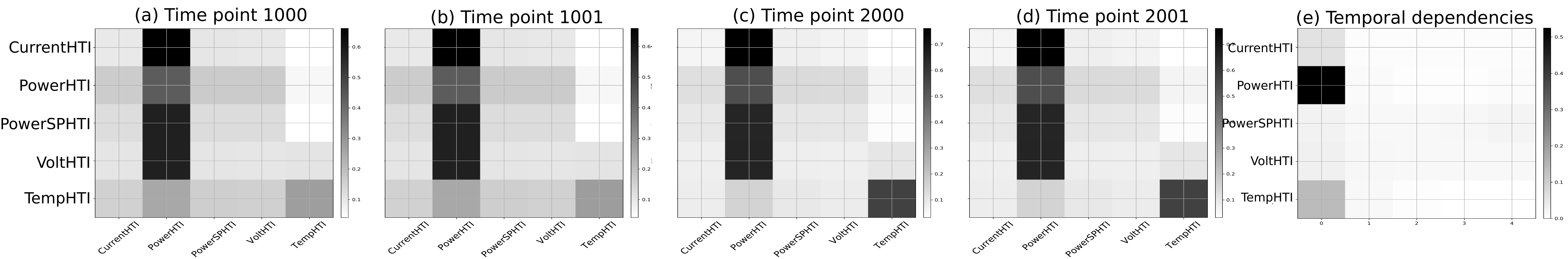}
\vspace*{-0.7cm}
\caption{Manufacturing Data: Common inter-dependencies are shown in the first four plots (at 4 different timestamps) while temporal lagged dependencies are summarized in 5th plot.}
\label{MNF-SA}
\end{figure*}


\begin{figure}[!h]
\hspace*{-1cm}
\includegraphics[width=9cm]{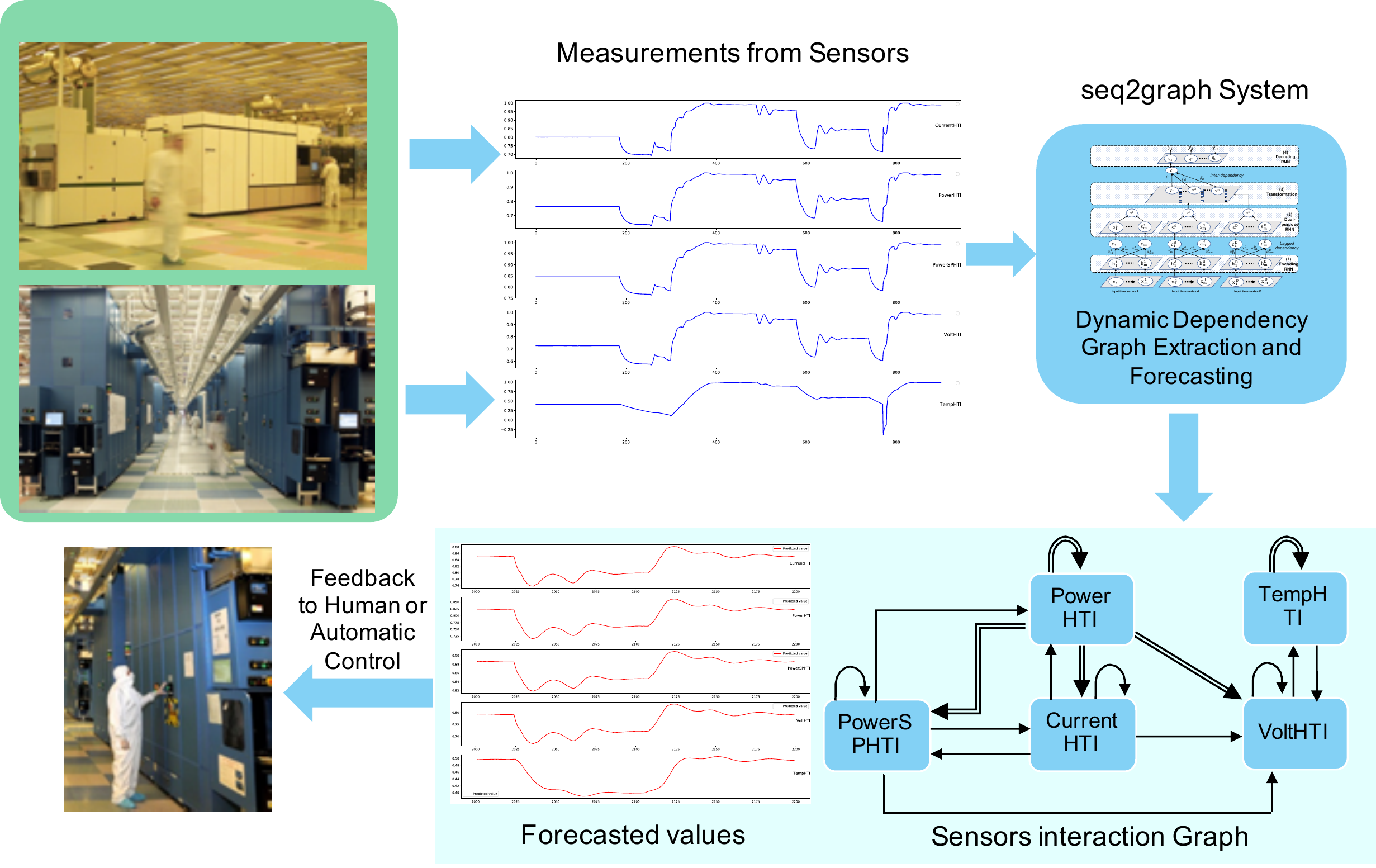}
\vspace*{-0.3cm}
\caption[Caption]{Application of our model to manufacturing data: our model analyzes sensor readings from different components of manufacturing plant and discovers dynamic dependencies among plant's components. It also forecasts their future behaviors and patterns which are provided as feedback to human or can be used to automatically take some action such as shutdown for safety reasons\footnotemark.
}
\label{ManufacturingPlant}
\end{figure}


For Manufacturing data, we show the application of our model in Fig.\ref{ManufacturingPlant}, and in Fig.\ref{MNF-SA} we present some typical dependencies (with milder changes). The probed time points at which these dependencies were investigated are shown on top of each plot. As observed, there are strong dependencies of all time series (except TempHTI) on the PowerHTI, which can be justified by the power law in electronics. However, future values of TempHTI are mostly dependent on its historical values, which shows its nature of being less variant and slowly varied as compared to other time series in the system. These dependencies are timely converted into directed weighted graphs in our testbed model as the one illustrated at the bottom-right graph in Fig.\ref{ManufacturingPlant}. The common time lagged dependencies are summarized in the last plot of Fig.\ref{MNF-SA}, revealing predictive information are mostly located at the latest time points of all time series, except with PowerSPHTI and VoltHTI whose further lagged time points also show impact. 

Recall that we do not have the ground truth of dependencies among component time series at any given time point of these real-world multivariate time series. Nonetheless, as an attempt to verify our dependency findings above, we select the Cloud Service Data and further perform the Granger causality tests (based on F-statistic)\cite{eichler2007granger,ACML2017}  of all other time series toward predicting 2 principle series of mem-util and cpu-util. The results are reported in Table~\ref{tab:Granger-IBM}. It is clearly observed that the P-value of pkg-rx is extremely small as compared to that of cpu-util toward predicting the mem-util series (top table), or that of mem-util toward predicting cpu-util (bottom table). This means we are more confident in rejecting the null hypothesis that pkg-rx does not Granger-cause mem-util (cpu-util), but less confident in rejecting the hypothesis that cpu-util (mem-util) does not Granger-cause mem-util (cpu-util). These statistical tests confirm the findings of our model \verb"seq2graph" above which shows the strong dependencies of both cpu-util and mem-util on pkg-rx, but less dependencies between cpu-util and mem-util in common situations (e.g. before time point 208 or after time point 214 in Fig.\ref{CloudServiceTS}). Note that, these pairwise tests also showed that pkg-tx may contain information toward predicting mem-util and cpu-util based on its low P-values. However, \verb"seq2graph" relies mainly on pkg-rx, rather than both pkg-rx and pkg-tx, in order to avoid redundant information, i.e. a new variable is included only if it brings additional information~\cite{guyon2003introduction}. 

It is worth mentioning that Granger-causality is only able to validate the existence of dependencies \textit{dominantly} appeared in the time series. It is, however, unable to detect the non-trivial dynamic changes in the dependencies of time series as demonstrated and explained in Fig.\ref{IBM-cloud-SA}, which are timely discovered by our learning network \verb"seq2graph".

 


\begin{table}[htbp]
  \centering
  \caption{Granger causality tests toward mem-util and cpu-util series in the Cloud Service Data.}
    \begin{tabular}{lrr}
    \textit{mem-util:} &       &  \\
    \midrule
    Ind. Time series & \multicolumn{1}{l}{F-statistic} & \multicolumn{1}{l}{P-value} \\
    \midrule
    pkg-rx & 147.1 & 9.20E-122 \\
    pkg-tx & 27.3  & 1.66E-22 \\
    cpu-util & 8.6   & 5.40E-07 \\
    \midrule
    &       &  \\
    \textit{cpu-util:} &       &  \\
    \midrule
    Ind. Time series & \multicolumn{1}{l}{F-statistic} & \multicolumn{1}{l}{P-value} \\
    \midrule
    pkg-rx & 1763.7 & 0 \\
    pkg-tx & 431.1 & 0 \\
    mem-util & 2.78  & 0.025 \\
    \bottomrule
    \end{tabular}%
  \label{tab:Granger-IBM}%
  \vspace*{-0.3cm}
\end{table}%

\footnotetext{Plant pictures source https://www.zdnet.com/pictures/inside-ibms-300mm-chip-fab-photos/ }
\paragraph{Prediction Accuracy:} Table~\ref{tbl-prediction-accuracy} shows forecasting accuracy comparison among all models on the Cloud Service Data, and Manufacturing Data. Although there are no big gaps among all examined methods, we still observe better performance for RNN models over the \verb"VAR" model in both datasets. Compared to the \verb"RNN-residual" and \verb"RNN-vanilla", our model performs more accurate on the Cloud Service Data, especially on the two principle time series mem-util and cpu-util. 



\section{Conclusions}

In this paper, we present a novel neural network model called \verb"seq2graph" that uses multiple layers of customized recurrent neural networks (RNNs) for discovering both time lagged behaviors as well as inter-timeseries dependencies during the \textit{inference} phase, which contrasts it to majority of existing studies that focus only on the forecasting problem. We extend the conventional RNNs and introduce a novel network component of dual-purpose RNN that decodes information in the temporal domain to discover time lagged dependencies within each time series, and encodes them into a set of feature vectors which, collected from all component time series, to form the informative inputs to discover inter-dependencies. All components in \verb"seq2graph" are jointly trained which allows the improvement in learning one type of dependencies immediately impact the learning of the other one, resulting in the overall accurate dependencies discovery. We empirically evaluate our model on non-linear, volatile synthetic and real-world datasets. The experimental results show that \verb"seq2graph" successfully discovers time series dependencies, uncovers the dynamic insights of the multivariate time series (which are missed by other established models), and provides highly accurate forecasts. 

In future, we aim to explore wider applications of our model and scale it to larger numbers of time series. Though several ablation studies showed that omitting either of components in our RNNs model did not lead to the right discovery of time lagged and inter-dependencies, it is worth to further investigate this issue with other successful network structures such as the convolution neural networks (CNNs) or the recently developed transformer network~\cite{vaswani2017attention}. We consider these as potential research directions.

\bibliographystyle{ACM-Reference-Format}
\bibliography{paper} 

\end{document}